%% file: WSDM-2019-DyGNN.tex
\documentclass[sigconf]{acmart}
\usepackage{cancel}

\usepackage{booktabs} 
\usepackage{multirow}
\usepackage{subfigure}
\usepackage{balance}
\setcopyright{rightsretained}

\acmDOI{10.475/123_4}

\acmISBN{123-4567-24-567/08/06}

\acmConference[WOODSTOCK'97]{ACM Woodstock conference}{July 1997}{El
  Paso, Texas USA}
\acmYear{1997}
\copyrightyear{2016}

\acmArticle{4}
\acmPrice{15.00}

\editor{Jennifer B. Sartor}
\editor{Theo D'Hondt}
\editor{Wolfgang De Meuter}

\begin{document}
\title{Streaming Graph Neural Networks}

\author{Yao Ma}
\affiliation{%
	\institution{Michigan State University}
}
\email{mayao4@msu.edu}

\author{Ziyi Guo}
\affiliation{%
	\institution{JD.com}
}
\email{guoziyi@jd.com}

\author{Zhaochun Ren}
\affiliation{%
	\institution{JD.com}
}
\email{renzhaocun@jd.com}

\author{Eric Zhao}
\affiliation{%
	\institution{JD.com}
}
\email{ericzhao@jd.com}

\author{Jiliang Tang}
\affiliation{%
	\institution{Michigan State University}
}
\email{tangjili@msu.edu}
\author{Dawei Yin}
\affiliation{%
	\institution{JD.com}
}
\email{yindawei@acm.org}

\begin{abstract}  
Graphs are essential representations of many real-world data such as social networks. Recent years have witnessed the increasing efforts made to extend the neural network models to graph-structured data. These methods, which are usually known as the graph neural networks, have been applied to advance many graphs related tasks such as reasoning dynamics of the physical system, graph classification, and node classification. Most of the existing graph neural network models have been designed for static graphs, while many real-world graphs are inherently dynamic. For example, social networks are naturally evolving as new users joining and new relations being created. Current graph neural network models cannot utilize the dynamic information in dynamic graphs. However, the dynamic information has been proven to enhance the performance of many graph analytic tasks such as community detection and link prediction. Hence, it is necessary to design dedicated graph neural networks for dynamic graphs. In this paper, we propose DGNN, a new {\bf D}ynamic {\bf G}raph {\bf N}eural {\bf N}etwork model, which can model the dynamic information as the graph evolving. In particular, the proposed framework can keep updating node information by capturing the sequential information of edges (interactions), the time intervals between edges and information propagation coherently. Experimental results on various dynamic graphs demonstrate the effectiveness of the proposed framework.

\end{abstract}

%
%

\maketitle

\input{sections/introduction}
\input{sections/model}
\input{sections/experiments}

\input{sections/related_work}
\input{sections/conclusions}

\bibliographystyle{ACM-Reference-Format}
\balance
\bibliography{reference/sample-bibliography}

\end{document}

%% file: sections/introduction.tex
\section{Introduction}\label{sec:introduction}

A graph describes a set of objects and their pairwise relations. Many real-life data such as social networks, transportation networks and e-commerce user-item graphs can naturally be represented in the form of graphs. Recent years have witnessed increasing efforts to generalize neural network models to graphs. These neural network models that operate on graphs are known as graph neural networks~\cite{gori2005new,scarselli2009graph}. Graph neural networks have been applied to perform the reasoning of dynamics of physical systems~\cite{battaglia2016interaction,chang2016compositional,sanchez2018graph}. Graph convolutional neural networks, which extend the convolutional neural networks to graph structure data, have been shown to improve the performance of graph classification~\cite{defferrard2016convolutional,bruna2013spectral} and node-level semi-supervised classification~\cite{kipf2016semi,hamilton2017inductive}. A general framework of graph neural network is proposed in~\cite{battaglia2018relational}. 

Most of these aforementioned neural network models have been designed for static graphs. Graphs in many real-world applications are inherently dynamic. For example, new users will join a social network and users in the social network will create new relations, users in e-commerce platform continue interacting with new items, and new connections are established in a communication network over time. To apply existing graph neural network models to dynamic graphs, we need to completely ignore their evolving structures by treating them as static graphs. However, the dynamic information has been proven to boost a variety of graph analytic tasks such as community detection~\cite{lin2008facetnet}, link prediction~\cite{goyal2018dyngem,li2018streaming} and network embedding~\cite{goyal2018dyngem,li2018streaming}. Therefore, it has great potential to advance graph neural networks by considering the dynamic nature of graphs, which calls for dedicated efforts. 

Meanwhile, designing graph neural networks for dynamic graphs faces tremendous challenges. From the global perspective, structures of dynamic graphs continue evolving since new nodes and edges are constantly introduced. It is necessary to capture the evolving structures for graph neural networks. From the local perspective, a node can keep establishing new edges, the establishing order of these edges is important to understand the node properties. For example, in the e-commerce user-item graph, new interactions are more likely to represent the users' latest preferences. Moreover, the introduction of a new edge (interaction) would affect the properties of the node. It is necessary to keep the node information updated once a new interaction happened. In addition, these edges are unevenly introduced, i.e., the distribution of these edges in the time-line is uneven. For example, a user in social networks could create edges very frequently in certain periods while only establishing a few edges in others. The time intervals between interactions for a specific node can vary dramatically. It is important to consider these time intervals and the major reasons are two-fold. First, the time interval between interactions of specific node can impact our strategy to update the node information. For example, if a new interaction is distant from its previous interaction, we should focus more on the new interaction since the node properties could change. Second, a new interaction can not only affect the two nodes directly involved in the interaction, but also can influence other nodes that are ``close'' to the two nodes; and the time interval can impact our strategy to propagate the interaction information to the influenced nodes. For example, if the new interaction is distant from the latest interaction between the node and an influenced node, the effect of the new interaction on the influenced node could be little. 

\begin{figure*}
	\centering
	\includegraphics[scale = 0.75]{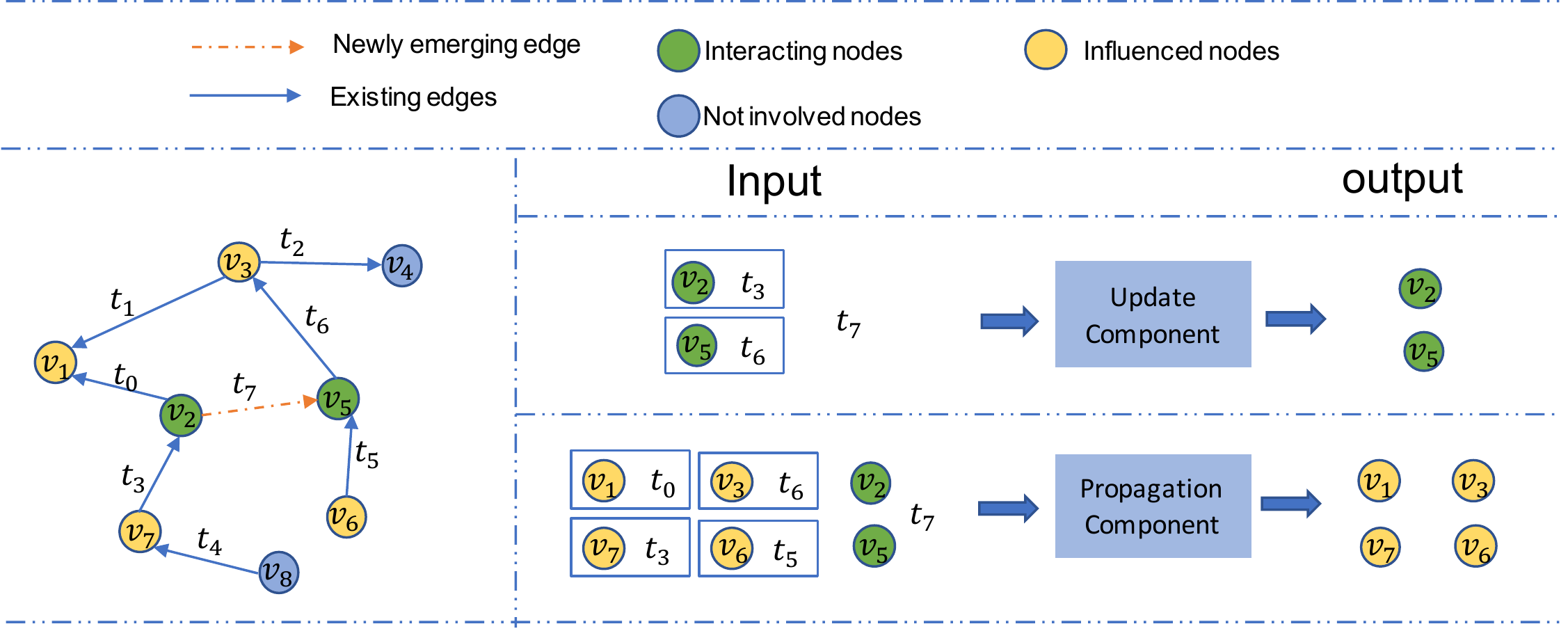}
	\caption{An overview of DGNN when a new interaction happened at time $t_7$ from $v_2$ to $v_5$. The two interacting nodes are $v_2$ and $v_5$. The nodes $\{v_1,v_3,v_6,v_7\}$ are assumed to be the influenced nodes.  }
		\label{fig:overall_framework}
\end{figure*}

In this paper, we embrace the opportunities and challenges to study graph neural networks for dynamic graphs. In essence, we aim to answer the following questions -- 1) how to constantly keep the node information updated when new interactions happen; 2) how to propagate the interaction information to the influenced nodes; and 3) how to incorporate time interval between interactions during update and propagation. We propose a dynamic graph neural networks (DGNN) to answer aforementioned three questions simultaneously. Our contributions can be summarized as follows:
\begin{itemize}
	\item We provide a principled approach for the node information update and propagation when new edges are introduced; 
	\item We propose a novel graph neural network for dynamic graphs (DGNN), which models establishing orders and time intervals of edges into a coherent framework; and 
	\item We demonstrate the effectiveness of the proposed model with several graph related tasks on various real-world dynamic graphs. 
\end{itemize}

The rest of this paper is organized as follows. In Section 2, we introduce the proposed framework with details about its update and propagation components and the approaches to learn model parameters. In Section 3, we present experimental results in two graph mining tasks including link prediction and node classification. We review related work in Section 4 and finally we conclude our work with future work in Section 5. 

%% file: sections/model.tex
\section{The proposed framework}

In this section, we introduce the graph neural network framework designed for dynamic networks. We first provide an overview about the model and then describe the components of the framework in details. Before that, we first introduce notations and definitions we will use in this work. 

A dynamic graph consists of a set of nodes and we assume that there are $N$ nodes $\mathcal{V} = \{v_1,v_2,\dots,v_N\}$ introduced until our latest observation about the graph. The graph evolves when new edges and nodes emerge. An example of a dynamic graph is shown in the left side of Figure~\ref{fig:overall_framework}, where there are $8$ nodes and $8$ interactions (edges) emerge from time $t_0$ to $t_7$. Note that, in this work, we only consider the emerging of new edges and nodes while leaving the deletion of existing edges and nodes as one future work. A directed edge $e$ can be represented as $(v_{s},v_{g},t)$ describing an interaction from $v_{s}$ to $v_{g}$ at time $t$. For example, the interaction happened at $t_0$ in Figure~\ref{fig:overall_framework} can be denoted as $\{v_2,v_1,t_0\}$. For convenience, we call the two nodes involved in the interaction as the ``interacting nodes''. As mentioned before, a new interaction can not only affect the two interacting nodes but also can influence other nodes that are ``close'' to the interacting nodes, which we call as the ``influenced nodes''. Thus, we need to update the information of this new interaction to the two interacting nodes and also propagate this information to the ``influenced nodes''.

To achieve this goal, a dynamic graph neural network (DGNN) is introduced and an overview about DGNN framework is demonstrated in Figure~\ref{fig:overall_framework}, which consists of two major components: 1) the update component and 2) the propagation component. We briefly describe the operations of the two components when introducing a new interaction $\{v_s,v_g,t\}$. The update component involves node $v_s$, node $v_g$ and updates the interaction information to both of them. For example, in Figure~\ref{fig:overall_framework}, a new interaction $\{v_2,v_5,t_7\}$ happened at $t_7$, the two interacting nodes being involved in the update component are $v_2$ and $v_5$. The propagate component involves the two interacting nodes $v_s, v_g$ and the ``influenced nodes'' as it propagates the information of the interaction $\{v_s,v_g,t\}$ to the ``influenced nodes''. The ``influenced nodes'' can be defined in different ways, which we will discuss in later subsections. In Figure~\ref{fig:overall_framework}, we define the ``influenced nodes'' as all the nodes that have interacted with the two ``interacting nodes'', which includes $\{v_1,v_7\}$---the 1-hop ``neighors'' of $v_2$, and $\{v_3,v_6 \}$--- the 1-hop ``neighbors'' of $v_6$. Next, we detail each component.

\subsection{The update component}
\begin{figure}[!h]
	\centering
	\includegraphics[scale = 0.28]{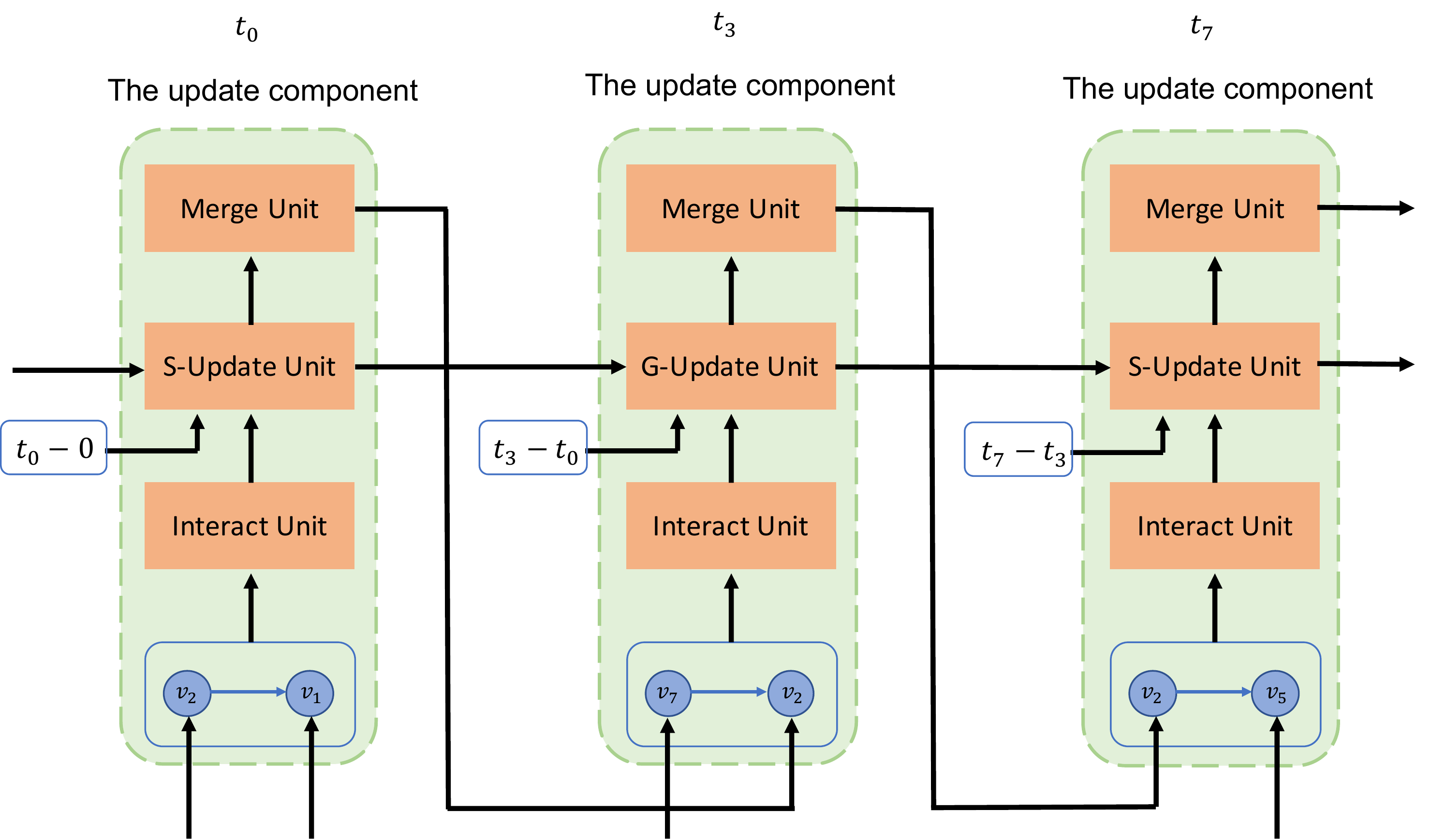}
	\caption{The operations of update components with the focus on node $v_2$ and all its interactions}
	\label{fig:update_components}
\end{figure}

In this subsection, we discuss the update component for the interacting nodes. We first give an overview of the operations of the update component with the focus on a single node $v_2$ of the dynamic graph illustrated in the left of Figure~\ref{fig:overall_framework}). There are three interactions involving node $v_2$, $\{v_2,v_1,t_0\}$, $\{v_7,v_2,t_3\}$ and $\{v_2,v_5,t_7\}$. It is natural that interactions between nodes will affect the properties of the nodes. For example, as suggested by homophily, users with similar interests are likely to create connections in social networks~\cite{mcpherson2001birds}. Thus, the update components should update the interaction information to the two interacting nodes. As shown in Figure~\ref{fig:update_components}, there are three update components, processing the three interactions involving node $v_2$ for node $v_2$. Each of the update components takes an interaction as input and update the interaction information to node $v_2$. Note that we only show the update component for node $v_2$ in Figure~\ref{fig:update_components}, while there is also another update component to update the interaction information to the other interacting node for each interaction. Furthermore, the order of the interactions is also important to understand the nodes' property. For example, in the e-commerce user-item graph, user's latest preference can be better captured by the recent interactions than the old ones. Thus, it is important to capture the order information. It is natural to view the interactions (involving the same node) as a ``sequence'' and recurrently apply the update component to the interactions. Note that, although the interactions can be viewed as a ``sequence'', we do not need to store all the information of this ``sequence''. We only store most recent information of the nodes. As shown in Figure~\ref{fig:update_components}, the three update components are connected in the sense that the next component takes the output of the previous component as input. Hence, we model the update component based on the long-short term memory (LSTM) unit~\cite{hochreiter1997long}. As discussed before, the time interval information is also important, thus, we also incorporate it into the update component. As shown in Figure~\ref{fig:update_components}, a single update component consists of three units -- the interact unit, the update unit and the merge unit. Next we describe these three units in details. 

Before proceeding to the details of the units, we first introduce the information we store for each node $v$. Note that in a directed graph, a node could play the roles of both source node and target node. Thus, we introduce two sets of different cell memories and hidden states for the two roles of each node, respectively. The cell memory and hidden state for the source role of node $v$ right before time $t$ are denoted as $C^s_c(t-)$ and $h^s_v(t-)$ respectively, while the cell memory and hidden state for the target role of node $v$ right before time $t$ are denoted as $C^g_c(t-)$ and $h^g_v(t-)$ separately. Here, the notation $t-$ means the time that is infinitely close to $t$, but prior to $t$, such that all the interactions before time $t$ have been processed. For example, in Figure~\ref{fig:update_components}, at time $t_7$, for node $v_2$, $C^s_c(t_7-)$ is, in fact, equal to $C^s_c(t_3)$. Note that we do not consider the propagation component now, for the purpose of illustrating the update component. The source and target hidden states are merged with the merge unit, to generate the general features $u_v(t-)$ of the node $v$, which describes the general property of node $v$. These cell memories $C^s_c(t-)$,  $C^g_c(t-)$, hidden states $h^s_v(t-)$, $h^g_v(t-)$ and general features $u_v(t-)$ are the information stored for each node $v$ and needed to be updated when new interaction happens. For example, Figure~\ref{fig:update_component} shows the operations of two update components performing update for node $v_2$ and $v_5$ when the interaction $\{v_2,v_5,t_7\}$ happens. The information stored for the two nodes right before time $t_7$ is shown in Figure~\ref{fig:update_components}~(a).

\subsubsection{The interact unit}
\begin{figure*}[!h]
	\centering
	\includegraphics[scale = 0.67]{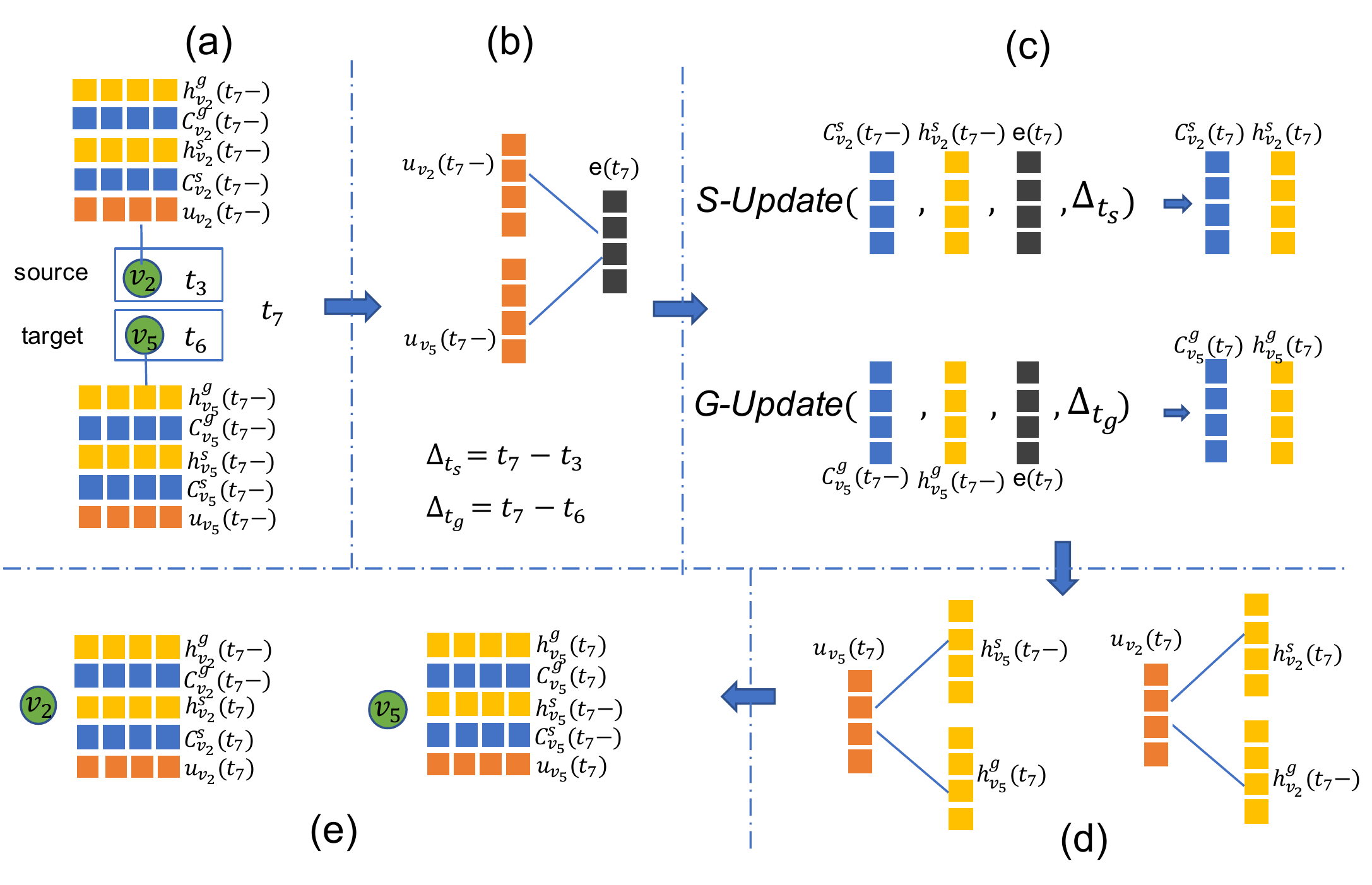}
	\caption{An example to illustrate an overview about the update components when an interaction ${v_2,v_5,t_7}$ happened.}
	\label{fig:update_component}
\end{figure*}

The interact unit is designed to generate the interaction information for $\{v_s,v_g,t\}$ from node information. The generated interaction information is later used as the input of the update unit. We model the interact unit using a deep feed-forward neural network and the formulation is as follows:
\begin{align}
e(t) = act(W_1 \cdot u_{v_s}(t-) +  W_2 \cdot u_{v_g}(t-) + b_e)
\end{align}
where $u_{v_s}(t-)$ and $u_{v_g}(t-)$ are the general features of the nodes $v_s$ and $v_g$ right before time $t$. $W_1$, $W_2$ and $b_e$ are the parameters of the neural network and $act(\cdot)$ is an activation function such as sigmoid or tanh. The output $e(t)$ contains the information of the interaction $\{v_s,v_g,t\}$. As an example, Figure~\ref{fig:update_component}~(b) shows how the interact unit works for the interaction $\{v_2,v_7,t_7\}$. 

\subsubsection{The update unit}
As mentioned before, the interactions (involving the same node) can be viewed as a ``sequence''. The information of this node gradually evolves as these interactions happens sequentially. Thus, to capture these interaction information for this node, we recurrently apply the update component to process the interaction information. The update unit is the part performing the operation to update the interaction information generated from the interact unit to the interacting nodes. Recall that the interactions do not emerge evenly in time. The time interval between interactions involving the same node can vary dramatically. The time interval impacts how the old information should be forgotten. It is intuitive that interactions happened in the far past should have less influence on the current information of node, thus they should be ``heavily" forgotten. On the other hand, recent interactions should have more importance on the current information of node. Thus, it is desired to incorporate the time interval into the update component. Hence, to build the update unit, we modify the LSTM unit as similar in~\cite{baytas2017patient} to incorporate the time interval information to control the ``magtitude'' of forgetting. 

\begin{figure}[!h]
	\centering
	\includegraphics[scale = 0.27]{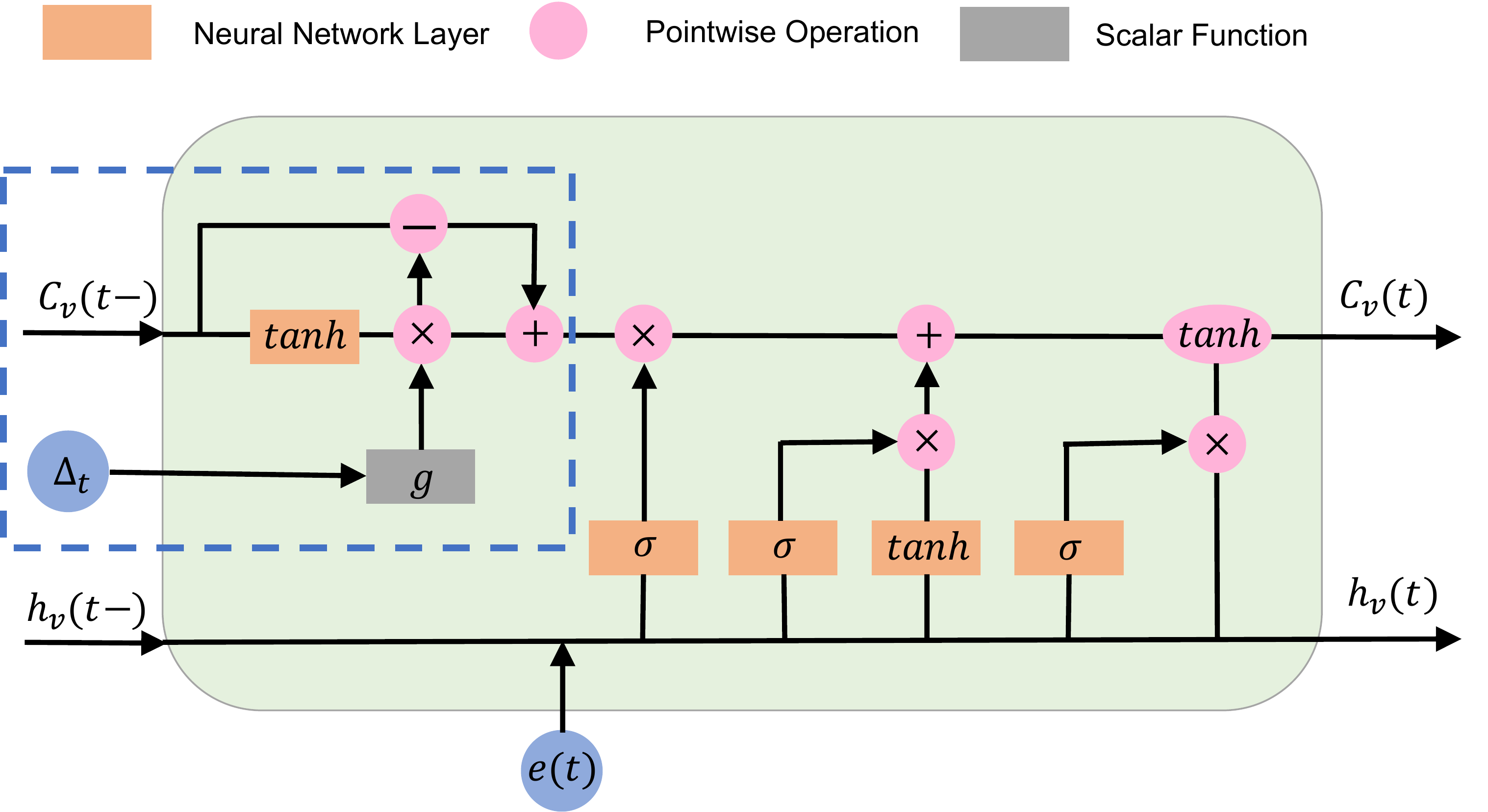}
	\caption{An illustration of the update unit}
	\label{fig:t_lstm_unit}
\end{figure}
An update unit is shown in Figure~\ref{fig:t_lstm_unit}, the input of this unit includes the most recent cell memory  $C_v(t-)$, hidden states $h_v(t-)$, the time interval $\Delta_t$ and the interaction information $e(t)$ calculated by the interact unit. The output of the update unit are the updated cell memory $C_v(t)$ and hidden state $h_v(t)$. Note that, for illustration purpose, we do not differentiate the source and target cell memory and hidden states in Figure~\ref{fig:t_lstm_unit}. In practice, we have two types of update units, the S-Update unit and the G-Update unit, which share the same structure but have different parameters. For an interaction $\{v_s,v_g,t\}$, we use the S-Update unit to update the information for the source node $v_s$ and use the G-Update unit to update the information for the target node $v_g$. The update unit is based on an LSTM unit, the only difference between the update unit and a standard LSTM unit is in the blue dashed box part of Figure~\ref{fig:t_lstm_unit}. The corresponding formulations for this part are as follows
\begin{align}
    	& C_v^I(t-1) = tanh(W_d\cdot C_v(t-1) +b_d)          \label{eq:start_of_update_unit} \\
	& \hat{C}_v^I(t-1) =  C_v^I(t-1) *g(\Delta_t)\\
	& C^T_v(t-1) = C_v(t-1) - C_v^I(t-1)\\
	&  C^*_v(t-1) = C^T_v(t-1) +  \hat{C}_v^I(t-1)
\end{align}
In this part, the old cell memory $C_v(t-1)$ is adjusted according to the time interval to generate the adjusted old cell memory $C^*_v(t-1)$. It is first decomposed to two components, the short term memory $C^I_v(t-1)$ and the long term memory $C^T_v(t-1)$, where $C^I_v(t-1)$ is generated by a neural network and the long term memory $C^T_v(t-1) = C_v(t-1) - C^I_v(t-1)$. The long term memory is kept untouched while the short term memory is discounted (forgotten) according to the time interval $\Delta_t$ between the events with a discount function $g$. The discount function $g$ is a decreasing function, which means the larger the time interval is, the less the short term memory is kept. Hence, we use this to model how we should forget the old information in our model. The discounted short term memory $\hat{C}^I_v(t-1)$ and the long term memory are then combined to generate the adjusted old cell memory $C^*_v(t-1) = \hat{C}^I_v(t-1) + C^T_v(t-1)$, which can be regarded as the output of the dashed box being input to the standard LSTM unit (the rest part of the update unit). The decomposition and recombination ensure that not the entire information of the old cell memory is lost during this procedure. The formulations of the rest part of the update unit, which are the same as a standard LSTM unit, are as follows
\begin{align}
    &f_t = \sigma(W_f \cdot  e(t) + U_f\cdot h_v(t-1) + b_f) \\
	&i_t = \sigma(W_i \cdot  e(t) + U_i\cdot h_v(t-1) + b_i) \\
	&o_t = \sigma(W_o \cdot  e(t) + U_o\cdot h_v(t-1) + b_o) \\
	&\tilde{ C}_v(t) = tanh(W_c \cdot  e(t) + U_c \cdot h_v(t-1) + b_c)\\
	&C_v(t)= f_t * C^*_v({t-1}) + i_t* \tilde{ C}_v(t)\\
	&h_v(t)  = o_t*tanh(C_v(t) ). \label{eq:end_of_the_update_unit}
\end{align}
For convenience, we summarize the procedure of the update unit in Figure~\ref{fig:t_lstm_unit} (eq.~\eqref{eq:start_of_update_unit} to eq.~\eqref{eq:end_of_the_update_unit}) as 
\begin{align}
	C_v(t), h_v(t) = \text{Update}(C_v(t-1),h_v(t-1),\Delta_t,e(t))
\end{align}
Examples of the operations of the update units are shown in Figure~\ref{fig:update_component}~(c), where, for interaction $\{v_2,v_7,t\}$ , we use the S-Update unit to update the information for the source node $v_2$ and use the G-Update unit to update the information for the target node $v_7$. Note that the S-Update unit only updates the source information (cell memory and hidden state) of the source node but keeps the target information of the source node untouched. Similarly, the G-Update unit only updates the target information of the target node but keeps the source information untouched. In Figure~\ref{fig:update_components}, for the convenience of illustration, we
``abuse'' the output of the S-Update unit and G-Update unit a little bit by considering the untouched target information of source node as part of the output of the S-Update unit and the untouched source information of target node as part of the output of the G-Update unit. 

\subsubsection{The merge unit}
The merge unit is to combine the source hidden state and target hidden state of a given node to generate the general features for this node. As we mentioned in last subsection, given an interaction $\{v_s,v_g,t\}$, the S-Update unit only updates the source information of the source node $v_s$ and the G-update only updates the target information of the target node $v_g$. Hence, for node $v_s$, we have $h^s_{v_s}(t)$ and $h^g_{v_s}(t-)$ as the output of the S-Update unit. The merge unit takes these two hidden states as input and generates new general features $u_{v_s}(t)$ for the node $v_s$ as follows:
\begin{align}
    u_{v_s}(t) = W^s\cdot h^s_{v_s} (t) + W^g\cdot h^g_{v_s}(t-) + b_u
\end{align}
Similarly, the merge unit generates the new general features $u_{v_g}(t)$ for node $v_g$ as follows:
\begin{align}
     u_{v_g}(t) = W^s\cdot h^s_{v_g} (t-) + W^g\cdot h^g_{v_g}(t) + b_u
\end{align}
The two merge units to generate new general features for node $v_2$ and $v_5$ after the interaction $\{v_2,v_5,t_7\}$ are shown in Figure~\ref{fig:update_component}~(d). 

Finally, the output of the update component is the updated information of the interacting nodes. For the source node $v_s$ of the interaction $\{v_s,v_g,t\}$, the updated information includes $C^s_{v_s}(t)$, $h^s_{v_s}(t), C^g_{v_s}(t-), h^g_{v_s}(t-)$ and $u_{v_s}(t)$. For the target node $v_g$, the updated information includes $C^s_{v_g}(t-), h^s_{v_g}(t-), C^g_{v_g}(t), h^g_{v_g}(t)$ and $u_{v_g}(t)$. The operations of the two update components for the interaction $\{v_2,v_5,t_7\}$ are shown in Figure~\ref{fig:update_component}.

\begin{figure*}
	\centering
	\includegraphics[scale = 0.7]{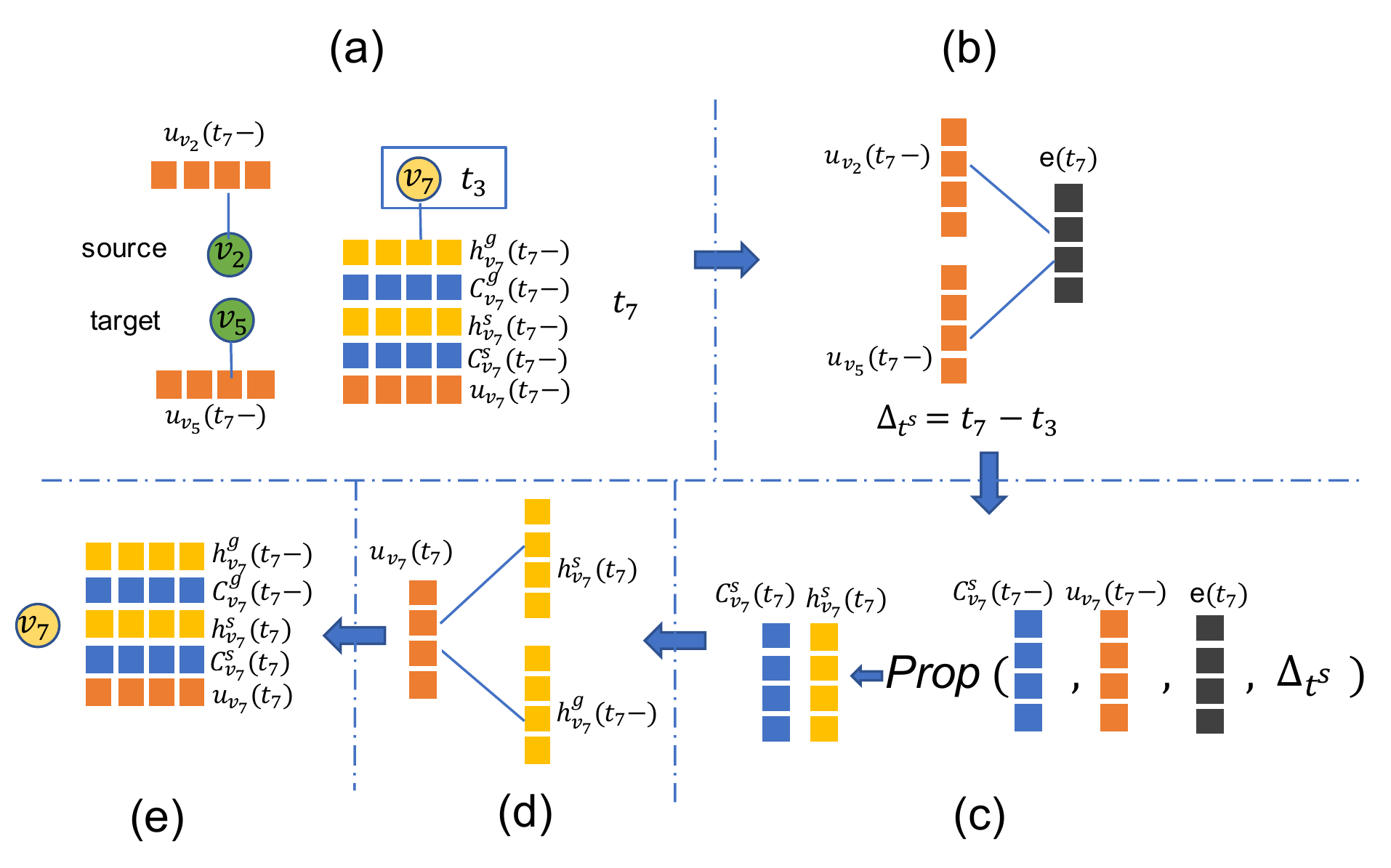}
	\caption{The propagation to the source neighbor $v_7$ of the source node $v_2$ when a new interaction $\{v_2,v_5,t_7\}$ happened. }
	\label{fig:prop_component}
\end{figure*}

\subsection{The propagation component}
In the previous section, we introduced the component to update the two interacting nodes when a new interaction happens. The update component only considers the two nodes directly affected by the new interaction. However, the newly emerging interaction changes the existing local structure of the graph. Thus, the interaction can influence some other nodes. In this work, we choose the current neighbors of the two ``interacting nodes'' as the ``influenced nodes''.  The major reasons are three-fold. First, as informed in mining streaming graphs, the impact of a new edge on the whole graph is often local~\cite{chang2017streaming}. Second, after we propagate information to the neighbors, the information will be further propagated, once the influenced nodes have interactions with other nodes. Third, we empirically found that when propagating more hops, the performance does not increase significantly or even decreases since we may also introduce noise during the propagation. To update the influenced nodes, the interaction information should be propagated to their cell memories. As the interaction does not directly influence the influenced nodes, we assume that the interaction does not disturb the history of the influenced nodes but only bring about new information. Thus, we do not need to decay or decrease the history information (cell memory) as what we do in the update component but only incrementally add new information to it. As similar with the intuition that older interactions should have less impact on the recent node information, an interaction should have less impact on the older influenced nodes. Thus, it is also desired to consider the time interval of the interactions in the propagation component. In addition, the influence can vary due to varied tie strengths (e.g., strong and weak ties are mixed together) ~\cite{xiang2010modeling}. Nodes are likely to influence others with strong ties than weak ties. Therefore, it is important to consider heterogeneous influence. With these intuitions, next we illustrate the operations of the propagation component. 

The propagation component consists of three units --  the interact unit, the prop unit and the merge unit. Note that the interact unit and the merge unit are the same as the ones in the update component. So, we mainly introduce the prop unit.

Let $\{v_s,v_g,t\}$ be the newly happened interaction, where $v_s$ is the source node and $v_g$ is the target node. The influenced nodes are the neighbors of these two nodes until time $t$, which can be denoted as $N(v_s)$ and $N(v_g)$. In a directed graph, we can further decompose the two sets of neighbors as $N(v_s)= N^s(v_s)\cup N^g(v_s)$ and $N(v_g)= N^s(v_g)\cup N^g(v_g)$, where $N^s(\cdot)$ denotes the set of source neighbors and $N^g(\cdot)$ denotes the set of target neighbors. Note that there are, in total, $4$ types of different prop units with the same structure but different parameters. They are the prop unit to propagate interaction information 1) from the source node $v_s$ to its source neighbors $N^s(v_s)$; 2) from the source node $v_s$ to its target neighbors $N^g(v_s)$; 3) from the target node $v_g$ to its source neighbors $N^s(v_g)$; and 4) from the target node $v_g$ to its target neighbors $N^g(v_g)$. We only describe one of them, from source node to its source neighbors, as the others have the same structure. For each node $v_x \in N^s(v_s)$, we propagate the interaction information to them with the following formulations:
{\footnotesize
\begin{align}
    & C^s_{v_x}(t) = C^s_{v_x}(t-) + f_a(u_{v_x}(t-), u_{v_s}(t-))\cdot g(\Delta_t^s) \cdot h(\Delta_t^s)\cdot \hat{W}_s^s\cdot e(t)\\
	& h^s_{v_x}(t) = \text{tanh}(C^s_{v_x}(t))
\end{align}
}
\noindent where $\Delta_t^s=t -t_x$ is the time interval between the current time $t$ and the last time $t_x$ when the node $v_x$ interacted with node $v_s$. $g(\Delta_t^s)$ is the same decay function as we defined for the update component. Intuitively, propagating the interaction information to ``extremely old neighbors'' may introduce noise. Hence, we introduce a function $h(\Delta_t^s)$ to filter some ``influenced nodes'' as defined as follows:

$$  h(\Delta_t^s)=\left\{
\begin{aligned}
 &1,  \quad \Delta_t^s \leq \tau,  \\
 &0,  \quad \text{otherwise}.
\end{aligned}
\right.
$$
where $\tau$ is a predefined threshold. This means if the time interval is too large ( $>\tau $), we will stop propagating information to such neighbors. {\it One advantage of this operation is to make the propagation step much more efficient}. We will demonstrate more details about this filtering step in the experiment section. $\hat{W}_s^s$ is a linear transformation to project the interaction information for propagating to source neighbors. We have different transformation matrix for the other types of prop units. The function $f_a(u_{v_x}(t_x-), u_{v_s}(t-))$ is an attention function to capture the tie strength between nodes $v_x$ and $v_s$ defined as:  
\begin{align}
f_a(u_{v_x}(t-), u_{v_s}(t-)) = \frac{\exp(u_{v_x}(t-)^T u_{v_s}(t-))}{\sum\limits_{v\in N^s(v_s)}\exp(u_{v}(t-)^T u_{v_s}(t-))}
\end{align} 
Figure~\ref{fig:prop_component} illustrates an example of propagating the interaction information to the source neighbor $v_7$ of the source node $v_2$ when an interaction $\{v_2,v_5,t_7\}$ happens. The prop unit is shown in Figure~\ref{fig:prop_component}~(c). Note that, for compactness of Figure~\ref{fig:prop_component}, we do not include the attention mechanism in it. The interact unit is shown in Figure~\ref{fig:prop_component}~(b), and the merge unit is shown in Figure~\ref{fig:prop_component}~(d).

\subsection{Parameter learning}
In this section, we introduce the parameter learning procedure of the dynamic graph neural network model. The proposed framework DGNN is general and can be utilized for a variety of network analytic tasks. Next we will use link prediction and node classification as examples to illustrate how to use DGNN for network analysis and its corresponding algorithm for parameter learning. 

\subsubsection{Parameter learning for link prediction}

To train the dynamic graph neural network model for the link prediction task, we design a specific training schedule. In DGNN, we only have one set of general features for each node, while each node can be either source node or target node. Thus, for the link prediction task, we first project the general features of the two interacting nodes to the corresponding role in the interaction with two projection matrix $P^s$ and $P^g$. We then adapt a widely used graph-based loss function with temporal information. For an interaction $(v_s,v_g,t)$, we project the most recent general features $u_{v_s}(t-)$, $u_{v_g}(t-)$ to $u^s_{v_s}(t-)$ and $u^g_{v_g}(t-)$ respectively as follows:
\begin{align*}
	&u^s_{v_s}(t-)  = P^s\cdot u_{v_s}(t-)\\
	&u^g_{v_g}(t-) = P^g \cdot u_{v_g}(t-)
\end{align*}

Then the probability of an interaction from $v_s$ to $v_g$ is modeled as $\sigma(u^s_{v_s}(t-)^T u^g_{v_g}(t-) )$ where $\sigma(\cdot)$ is the sigmod function. Eventually the loss can be represented as
\begin{align}
J( (v_s,v_g,t) ) =& -log(\sigma(u^s_{v_s}(t-)^T u^g_{v_g}(t-) )) \nonumber\\
& - Q\cdot \mathbb{E}_{v_n \sim P_n(v)}  log(\sigma(u^s_{v_s}(t-)^T u^g_{v_n}(t-) ))
\end{align}
where $Q$ is the number of negative samples and $P_n(v)$ is a negative sampling distribution. The total loss until time $T$ can be represented as
\begin{align}
	\sum\limits_{e\in\mathcal{E}(T)} J(e);
\end{align}
where $\mathcal{E}(T)$ denotes all the interactions until time $T$. 

We then adopt a mini-batch gradient descent method to optimize the loss function. Note that in our case, the mini-batches of edges are not randomly sampled from the entire set of edges but sequences from the interaction sequence maintaining the temporal order.  The loss of the mini-batch is calculated from all the interactions in the mini-batch. The negative sampling distribution $P_n{(v)}$ is a uniform distribution out of all the nodes involved in the mini-batch, which includes the interacting nodes and the influenced nodes of each interaction. 

\subsubsection{Learning parameters for node classification}
To train the dynamic graph neural network model for node classification, we adopt the cross entropy loss. For a node $v$ with general features $u_{v}(t)$ and label $y\in \{0,1\}^{N_c}$ immediately after time $t$, where $N_c$ is the number of classes, we first project $u_{v}(t)$ to $u^c_{v}(t)\in \mathcal{R}^{N_c \times 1}$. Then, the loss corresponding for the node $v$ at time $t$ is defined as
\begin{align*}
  J(v,t) = -	\sum\limits_{i=0}^{N_c-1}  y[i] \log\left (\frac{\exp(u^c_{v}(t))[i]}{\sum\limits_{j=0}^{N_c-1}\exp(u^c_{v}(t))[j]}\right);
\end{align*}
where $y[i]$ and $u^c_{v}(t)[i]$ denote the i-th element of $y$ and $u^c_v(t)$, respectively. 

The training schedule is semi-supervised, only some nodes are labeled but the unlabeled nodes are also involved in the update and propagation component of the dynamic graph neural network model. We adopt a similar mini-batch (of edges) procedure as that in link prediction. Let $T_m$ be the end time of the mini-batch, i.e. the time of the last interaction in the mini-batch. After the mini-batch of interactions is processed by the update and propagation components of DGNN, we collect all the nodes involved in the mini-batch and denote them as $\mathcal{V}_m$. Let $\mathcal{V}_{train}$ denote the set of all the nodes with labels. We then use $ \mathcal{V}_{m-train} = \mathcal{V}_m \cap \mathcal{V}_{train}$ as the training samples of this mini-batch. We form the loss function for this mini-batch as 
\begin{align}
	\sum\limits_{v\in \mathcal{V}_{m-train} } J(v,T_m).
\end{align}

%% file: sections/experiments.tex
\section{Experiments}
In this section, we perform two graph based tasks to demonstrate the effectiveness of the proposed dynamic graph network model. We first introduce three datasets we use in the experiments, then present the two tasks---link prediction and node classification---with experimental details and discussions and finally study the key model components of the proposed framework.  

\subsection{Datasets}
We conduct the experiments on the following three datasets. Some important statics of the three datasets can be found in Table~\ref{tab:statistics}.
\begin{table}[h]
	\begin{center}
		\caption{Statistics of datasets\label{tab:statistics}}
		\vspace*{-0.15in}
		\begin{tabular}{c |c|c |c} 
			\hline
			& UCI & DNC& Epinions \\ 
			\hline
			number of nodes &1,899 &2,029 &6,224\\
			\hline
			number of  edges &59,835& 39,264 & 19496\\
			\hline
			time duration &194 days  & 982 days &936 days\\
			\hline
			number of labels &$ \backslash$  & $\backslash$ &15\\
			\hline 
		\end{tabular}
	\end{center}
\end{table}
\begin{itemize}
	\item {\bf UCI}~\cite{kunegis2013konect} is a directed graph which denotes the message communications between the users of an online community of students from the University of California, Irvine. A node in this graph represents a user. There are edges between users if they have message communications where the time associated with each edge indicates when they communicated. In this dataset, the graph structure and edge creation time are available; hence we use this dataset to evaluate link prediction performance. 
		\item {\bf DNC}~\cite{kunegis2013konect} is a directed graph of email communications in the 2016 Democratic National Committee email leak. Nodes in this graph represents persons. A directed edge in this graph represents that an email is sent from one person to another. In this dataset, the graph structure and the time information when edges are established are available; thus we use this dataset for the link prediction task. 
	\item {\bf Epinions}~\cite{tang-etal12a} is a directed graph which denotes trust relations between users in the product review platform Epinions. A node in this graph represents a user. A directed edge represents a trust relation among users. We have $15$ labels in this dataset. The label of each user is assigned according to the category of the majority of the user's reviewed products. In this dataset, we have graph structure, edge creation time and node labels; therefore, we use this dataset for both link prediction and node classification tasks. 
\end{itemize}

\subsection{Link prediction}
In this section, we conduct the link prediction experiments to evaluate the performance of the DGNN model. We first introduce the baselines. We then describe the experimental setting of the link prediction task and the evaluation metrics. Finally, we present the experimental results with discussions. 

\subsubsection{Baselines}

We carefully choose representative baselines from two groups.  One group includes existing neural graph network models. The other contains state of art graph representation learning methods given their promising performance in link prediction. Details about baselines are introduced as follows: 
\begin{itemize}
	\item \textbf{GCN}~\cite{kipf2016semi} is a state of art graph convolutional network model, it tries to learn better node features by aggregating information from the node's neighbors. The method cannot use temporal information; thus we treat the dynamic graphs as static graphs for this method by ignoring the edge creation time information.
	\item \textbf{GraphSage}~\cite{hamilton2017inductive} also aggregates information from neighbors, but it samples neighbors instead of using all neighbors.  It cannot use temporal information neither; thus we treat the dynamic graphs as static graphs for this method similar to \textbf{GCN}. 
	\item \textbf{node2vec}~\cite{grover2016node2vec} is a state of art graph representation learning method. It utilizes random walk to capture the proximity in the network and maps all the nodes into a low-dimensional representation space which preserves the proximity. It cannot utilize the temporal information and we convert the dynamic graphs into static graphs for node2vec.
	\item \textbf{DynGEM}~\cite{goyal2018dyngem} is a graph representation learning method designed for dynamic graphs. However, it can only be applied to discrete time data with snapshots, thus in our experiments, we split each dataset into snapshots for this baseline.
	\item \textbf{CPTM}~\cite{dunlavy2011temporal} is a tensor-based model. It treats the dynamic graph as $3$-dimension tensor, where two dimensions describe the interactions of nodes and the third dimension is time. It decomposes the tensor to get the features of the nodes. It can only be applied to discrete time data with snapshots. Hence, as for DyGEM, we split each dataset into snapshots.
	\item \textbf{DANE}~\cite{li2017attributed} is a recent proposed eigendocompoation based node representation learning algorithm for attributed dynamic graphs. It updates the node representations over time by perturbation analysis of eigenvectors. It can only be applied to discrete time data with snapshots. Hence, as for DyGEM, we split each dataset into snapshots. Note that since the focus in this work is not attributed networks, we use a variant of \textbf{DANE}, which only considers the structural information of dynamic networks. 
	\item \textbf{DynamicTriad}~\cite{zhou2018dynamic} is a recent proposed node representation learning algorithm for dynamic graphs. As suggested by its name, it is based on modelling the triangle closure between snapshots of the dynamic graphs. It can only be applied to discrete time data with snapshots. Hence, as for DyGEM, we split each dataset into snapshots. 
\end{itemize} 

As we can see, our baselines include representative graph neural network models, i.e., \textbf{GCN} and \textbf{GraphSage}, one state of the art static node embedding method \textbf{node2vec}, three recent dynamic network embedding methods \textbf{DynGEM}, \textbf{DANE}, \textbf{DynamicTriad} and one traditional dynamic network embedding method \textbf{CPTM}. 

\subsubsection{Experimental setting}

In the link prediction task, we are given a fraction of interactions in the graph as the history and supposed to predict which new edges will emerge in the future. In this experiment, we use the first $80\%$ of the edges as the history (training set) to train the dynamic graph neural network model, $10\%$ of the edges as the validation set and the next $10\%$ edges as the testing set. All the baselines and our model return node features after training. We use the node features learned with the $80\%$ training set as the node features for link prediction. For each edge $(v_{s}, v_{g}, t)$ in the testing set, we first fix $v_{s}$ and replace $v_{g}$ with all nodes in the graph and then we use the cosine similarity to measure the similarity and rank the nodes. We then fix $v_{g}$ and replace $v_{s}$ with all the nodes in the graph and rank the nodes in a similar way. For all the models we tune the parameters on the validation set. For our model, to calculate the cosine similarity, we use the projected features for UCI and DNC dataset and use the original features for Epinions dataset according to the performance on the validation dataset. In the link prediction task, we randomly initialize the cell memories, hidden states and general features for all nodes. 

\subsubsection{Evaluation metrics}

We use two different metrics to evaluate the performance of the link prediction task. One of them is mean reciprocal rank (\textbf{MRR})~\cite{voorhees1999trec}, which is defined as
\begin{align}
\textbf{MRR} = \frac{1}{|H|} \sum\limits_{i=1}^{H}\frac{1}{rank_i}
\end{align}
where $H$ is the number of testing pairs. Note that one edge is corresponding to two testing pairs: one for the source node and the other one for the target node. $rank_i$ is the rank of the ground truth node out of all the nodes. The \textbf{MRR} metric calculates the mean of the reciprocal ranking of the ground truth nodes in the testing set. It is higher when there are more ground truth node ranked top out of all the nodes. 

The other metric we use is \textbf{Recall@k}, which is defined as:
\begin{align}
	\textbf{Recall@k} = \frac{1}{|H|} \sum\limits_i^{|H|}\mathbf{1}\{rank_i \leq k\}  
\end{align}
where $\mathbf{1}\{rank_i \leq k\} =1$ only when $rank_i\leq k$, otherwise $0$. The \textbf{recall@k} calculates how many of the ground truth nodes are ranked in top $k$ out of all the nodes in their own testing pairs. The larger it is, the better the performance is. In this work, we use \textbf{Recall@20} and \textbf{Recall@50}. 

\subsubsection{Experimental results}
In this section, we present the experimental results. The link prediction results on the three datasets are shown in Table~\ref{tab:link_prediction}. From results, we can make the following observations
\begin{itemize}
    \item DANE does not perform well as expected since it has been originally designed for attributed networks. 
	\item DynGEM and DynamicTriad outperforms node2vec in most cases. All the three methods are embedding algorithms -- node2vec is for static networks while DynGEM and DynamicTriad capture dynamics. These results suggest the importance of the dynamic information in graphs. 
	\item The proposed dynamic graph neural network model outperforms two representative existing GNNs, i.e., GCN and GraphSage. Our model is for dynamic networks while GCN and GraphSage ignore the dynamic information, which further support the importance to capture dynamics. 
    \item  The proposed model DGNN outperforms all the baselines in most of the cases on all the three datasets. DGNN provides model components to capture time interval, propagation and tie strength. In the following subsections, we will study the impact of these model components on the performance of the proposed framework. 
\end{itemize}

\begin{table*}
	\begin{center}	
		\caption{Performance comparison of link prediction.}
		\label{tab:link_prediction}
		\begin{tabular} { c| c c c| c c c| c c c }
			
			\hline 
			\multirow{ 2}{*}{Baselines} & \multicolumn{3}{c}{UCI} & \multicolumn{3}{c}{DNC} & \multicolumn{3}{c}{Epinions}\\
			&  MRR & Recall@20 & Recall@50   &  MRR & Recall@20 & Recall@50  &  MRR & Recall@20 & Recall@50  \\	
			\hline			
			DGNN & {\bf 0.0342}&   {\bf 0.1284} &  {\bf 0.2547} & {\bf 0.0536} &  0.1852 &  {\bf 0.3884}& {\bf 0.0204} &  {\bf0.0848} &  {\bf 0.1894}\\ [0.5ex]	
			GCN &  0.0138 &   0.0632&   0.1176 &  0.0447  &{\bf 0.2032}&  0.3291&  0.0045 &  0.0071 &  0.0119\\[0.5ex]	
			GraphSage & 0.0060&  0.0161  &  0.0578  &   0.0167 &  0.0576 &  0.1781&  0.0035 &  0.0072&   0.0108\\[0.5ex]
			node2vec&  0.0056 &  0.0184 &  0.0309  & 0.0202 &   0.0719&   0.178&  0.0135 &  0.0571 &  0.1240\\[0.5ex]
			DynGEM& 0.0146 &  0.0773  &   0.1455  &  0.0271 &  0.0971 &  0.2356 &  0.0150 &0.0657  & 0.1233 \\[0.5ex]
			CPTM & 0.0138 &  0.0921 &  0.1082  & 0.0109 &  0.0072 &   0.0108 &   0.0036&  0.0060 &  0.0125\\[0.5ex]
			DANE & 0.0040&  0.0110 &  0.0233  & 0.0128 &  0.0270 &   0.0432 &   0.0040&  0.0100 &  0.0120\\[0.5ex]
			DynamicTriad & 0.0150&  0.0610 &  0.1236  & 0.0146 & 0.0414 &0.0665   &   0.0170 &  0.0729 &  0.1629\\[0.5ex]
			\hline   
		\end{tabular}
	\end{center}
\end{table*}

\subsection{Node classification}
In this subsection, we conduct the node classification task to evaluate the performance of the dynamic graph neural network model. We first introduce the baselines. We then describe the experimental setting and the evaluation metrics. Finally, we present the experimental results. 

\subsubsection{Baselines}
The node classification task is a semi-supervised learning task, where some nodes are labeled and we aim to infer the labels of unlabeled nodes in the graph.   Therefore, we carefully choose two groups of baselines. One is about GNNs for semi-supervised learning including GCN and GraphSage. The other is traditional semi-supervised learning methods and we choose a start-of-the-art traditional semi-supervised method LP based on Label Propagation~\cite{zhu2003semi}. Note that for a fair comparison, we do not choose node embedding algorithms such as node2vec as baselines since they are designed under the unsupervised setting. 

\subsubsection{Experimental setting}
In the node classification task, we randomly sample a fraction of nodes and hide their labels. These nodes with labels hidden will be treated as validation and testing sets. The remaining nodes are treated as the training set. In this work, we randomly sample $20\%$ of all the nodes and hide their labels. We use $10\%$ of them as validation set and the other $10\%$ as testing set. For the rest $80\%$ of nodes with labels, we choose $x\%$ as labeled nodes and others as unlabeled nodes. In this experiment, we vary $x$ as $\{100, 80, 60\}$. We use $F_1$-micro and $F_1$-macro as the metrics to measure the performance of the node classification task. 

\subsubsection{Experimental results}
\begin{figure*}
	\begin{center}
		\subfigure[Epinions: $F_1$-macro]{\label{fig:f1_macro}\includegraphics[scale=0.5]{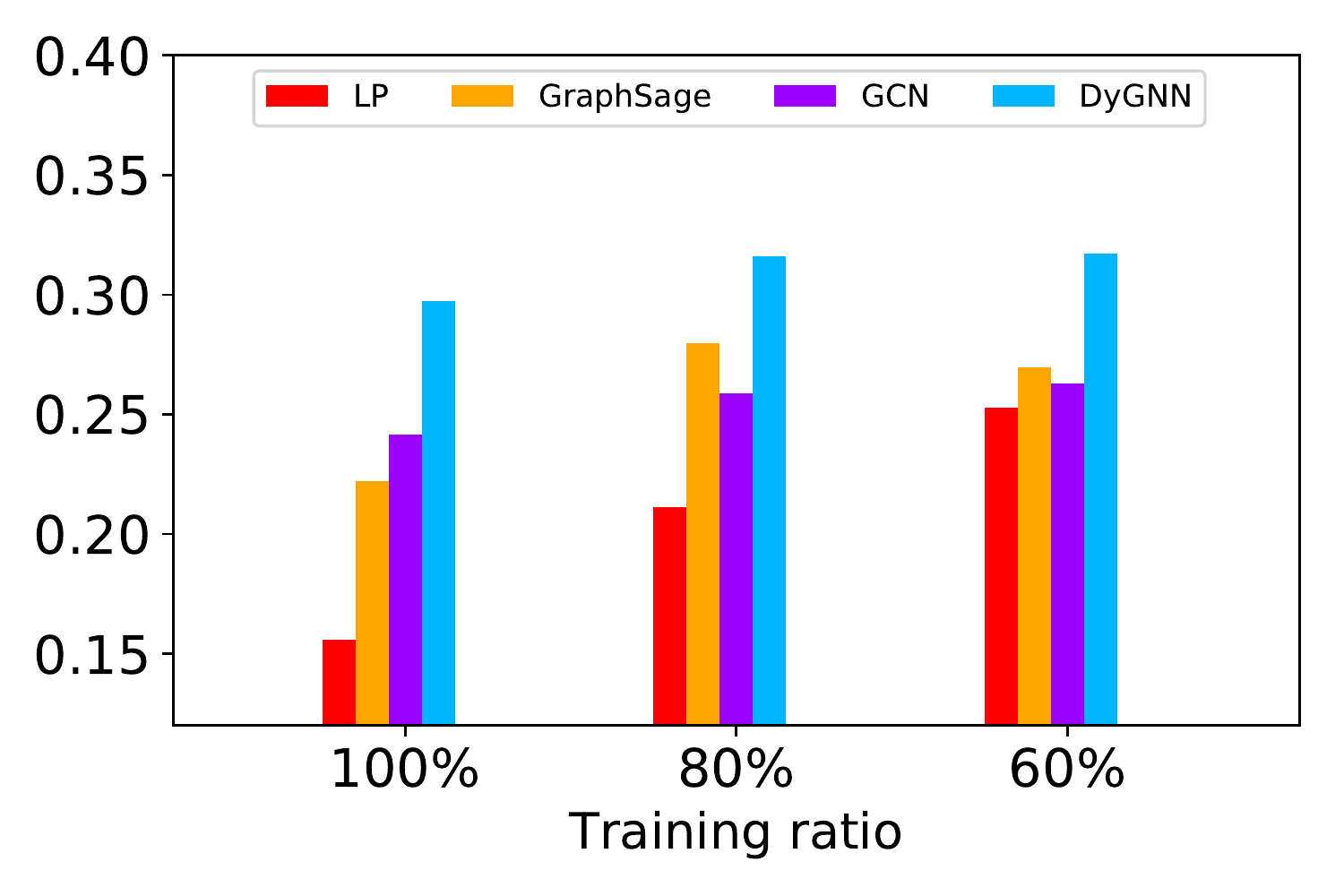}}
		\subfigure[Epinions: $F_1$-micro]{\label{fig:f1_micro}\includegraphics[scale=0.5]{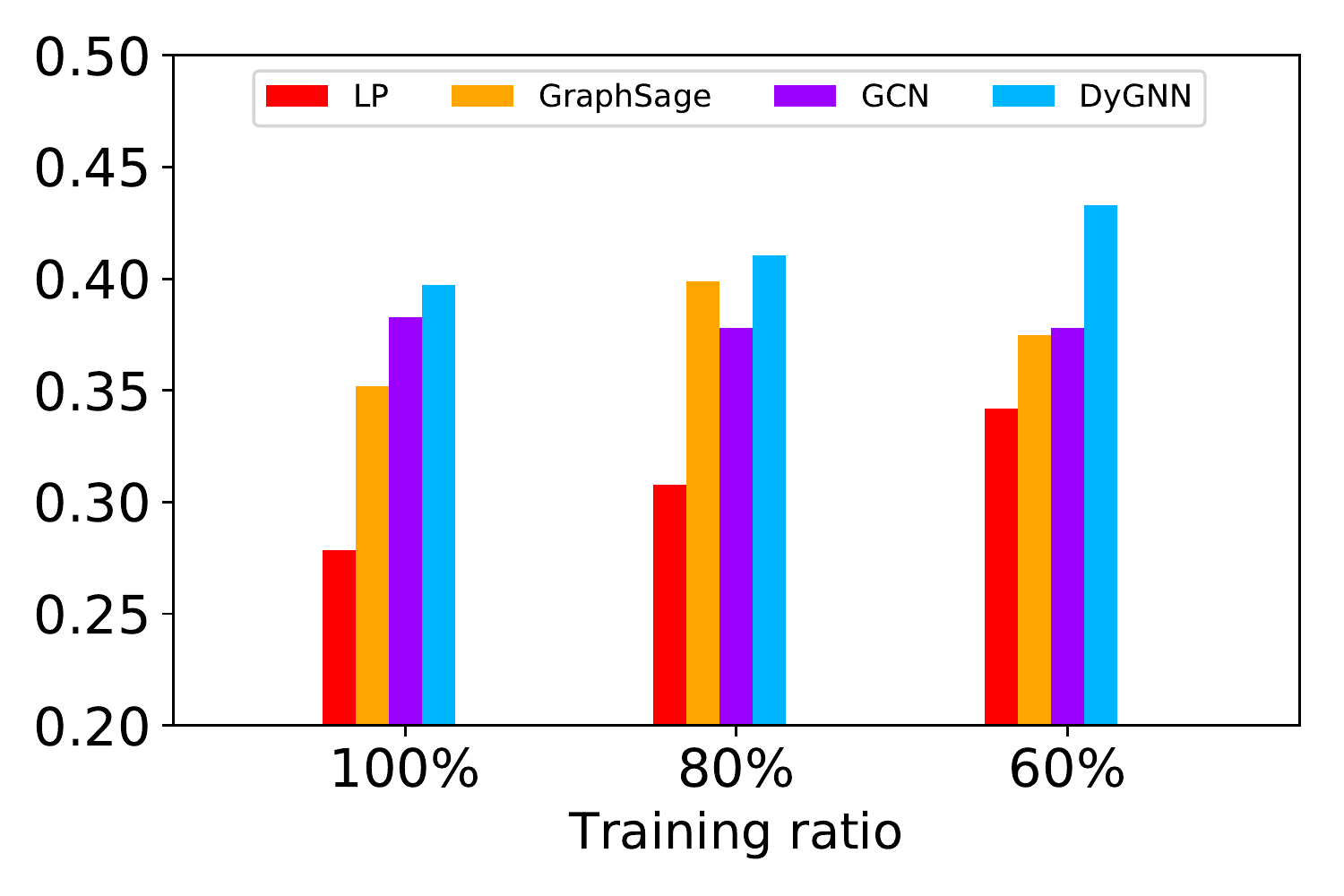}}
	\end{center}
	\caption{Performance Comparison of Node classification on Epinions dataset}
	\label{fig:nc_results}
\end{figure*}
Among three datasets, only Epinions dataset has label information. Hence we conduct the node classification task on it and the results are presented in Figure~\ref{fig:nc_results}.  We can make the following observations: 
\begin{itemize}
    \item With the increase of the number of labeled nodes, the classification performance tends to increase. 
	\item GraphSage, GCN and DGNN outperforms LP in all settings, which indicates the power of GNNs in semi-supervised learning. 
	\item DGNN outperforms GraphSage and GCN under all the three settings, which shows the importance of temporal information in node classification.
\end{itemize}

\subsection{Model Component analysis}

In the last two sections, we have demonstrated the effectiveness of the proposed framework in two graph mining tasks -- link prediction and node classification. In this subsection, we conduct experiments to understand the effect of the key components on our proposed model. More specifically, we form the following variants of our model by removing some components in the model:
\begin{itemize}
\item \textbf{DGNN-prop:} In this variant, we remove the entire propagation component from the model. This variant only does the update procedure when new edge emerges. 
\item \textbf{DGNN-ti:} In this variant, we do not use the time interval information in both update component and propagation component. Thus, we treat the interactions as a sequence with no temporal information. 
\item \textbf{DGNN-att:} In this variant, we remove the attention mechanism in the propagation component and consider equal influence. 
\end{itemize}

We will use the task of link prediction to illustrate the impact of model components. The performance of these variants on link prediction task are shown in Table~\ref{tab:variants_comparison}. As we can observe from the results, all the three components are important to our model, as removing them will reduce the performance of link prediction. Via this study, we can conclude that (1) it is necessary to propagate interaction information to influenced nodes; (2) it is important to consider the time interval information; and (3) capturing varied influence can improve the performance. 

\begin{table*}
	\begin{center}	
		\caption{Comparison of variants on the link prediction task.}
		\label{tab:variants_comparison}
		\begin{tabular} { c |c c c| c c c| c c c }
			
			\hline 
			\multirow{ 2}{*}{Baselines} & \multicolumn{3}{c}{UCI} & \multicolumn{3}{c}{DNC} & \multicolumn{3}{c}{Epinions}\\
			&  MRR & Recall@20 & Recall@50   &  MRR & Recall@20 & Recall@50  &  MRR & Recall@20 & Recall@50  \\	
			\hline			
			DGNN & 0.0342&   0.1284 &  0.2547 & 0.0536 &  0.185 &  0.3884& 0.0204 &   0.0848 &  0.1894\\ [0.5ex]	
			DGNN-prop &  0.0103 &   0.0444&   0.1087 &  0.0046  &  0&  0&  0.0171 &  0.0633 &  0.1514\\[0.5ex]	
			DGNN-ti & 0.0174&  0.0918  &  0.2118  &   0.0050 &  0 & 0.0054&  0.0157 &  0.0591&   0.1589\\[0.5ex]
			DGNN-att&  0.0200 &  0.0844 &  0.2235  & 0.0562 &   0.1547&   0.3219&  0.0177 &  0.0651 &  0.1655\\[0.5ex]
			\hline   
		\end{tabular}
	\end{center}
\end{table*}
\begin{figure}[!h]
	\centering
	\includegraphics[scale = 0.5]{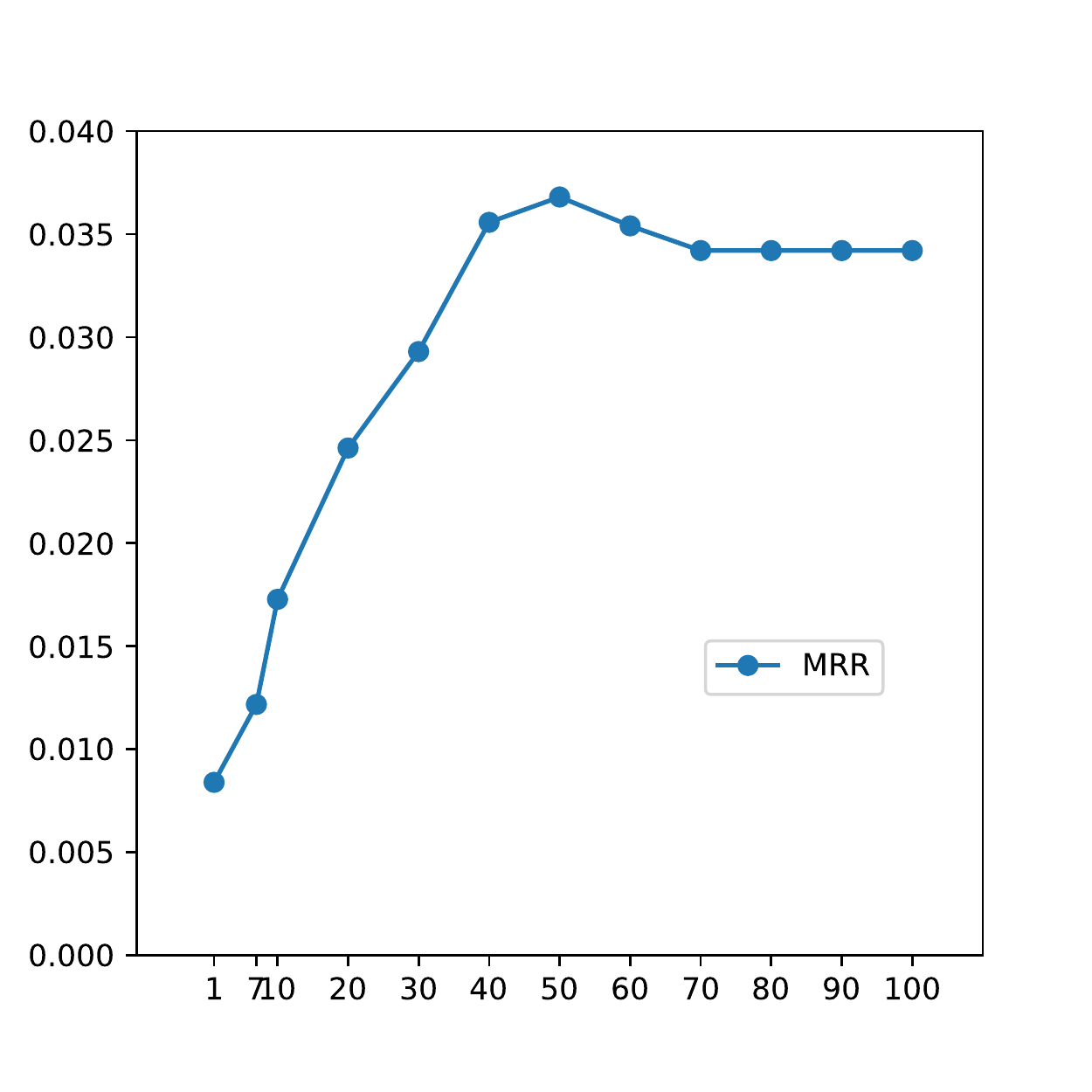}
	\vspace{-0.2in}
	\caption{Impact of $\tau$}
	\vspace{-0.3in}
	\label{fig:threshold}
\end{figure}

\subsection{Parameter Analysis}
The proposed framework introduces one parameter $\tau$ in the propagation component to filter some ``influenced nodes". In this subsection, we analyze how different values of $\tau$ in the propagation component affect the performance of the DGNN model. We perform the analysis for the link prediction task on the UCI dataset with the $MRR$ measure since we have similar observations with other settings and on other datasets. 

As shown in Table~\ref{tab:statistics}, the duration of this dataset is $194$ days and we set the threshold $\tau$ to $1,7$ days, and $10-100$ days with a step size of $10$. The performance in terms of $MRR$ is shown in Figure~\ref{fig:threshold}. The performance of DGNN first increases as the threshold $\tau$ gets larger. A large $\tau$ allows the interaction information to be propagated to more influenced nodes. After $\tau$ hits $50$, the performance becomes stable or even slightly decreases. These observations suggest that 1) the propagation procedure does help to broadcast necessary information to the ``influenced nodes'' as the performance first gets improved when the threshold increases; and 2) propagating the interaction information to ``extremely old neighbors'' may not be helpful or even may bring noise. These observations have practical significance since we can choose a proper $\tau$ in the propagation component, which can remarkably boost the efficiency of the proposed framework since we only need to perform the propagation with a small number of ``influenced nodes".

%% file: sections/related_work.tex
\section{Related work}
In this section, we briefly review two streams of research related to our work: graph neural networks and dynamic graph analysis.   

In recent years, many efforts have been made to extend deep neural network models to graph structured data. These neural network models that are applied to graphs are known as graph neural network models~\cite{gori2005new,scarselli2009graph}. They have been applied to various tasks in many areas. Various graph neural network models have been designed to reason dynamics of physical systems where previous states of the nodes are given as history to predict future states of the nodes~\cite{battaglia2016interaction,chang2016compositional,sanchez2018graph}. Neural message passing networks have been designed to predict the properties of molecules~\cite{gilmer2017neural}. Graph convolutional neural networks, which try to perform convolution operations on graph structure data, have been shown to advance many tasks such as graph classification~\cite{defferrard2016convolutional,bruna2013spectral}, node classification~\cite{kipf2016semi,hamilton2017inductive,velivckovic2017graph} and recommendation~\cite{ying2018graph}.
A comprehensive survey on graph convolutional neural networks can be found in~\cite{bronstein2017geometric}. A general framework of graph neural networks was proposed in~\cite{battaglia2018relational} recently.

Most of the current graph neural network models are designed for static graphs where nodes and edges are fixed. However, many real-world graphs are evolving. For example, social networks are naturally evolving as new nodes joining the graph and new edges being created. It has been of great interest to study the properties of dynamic graphs~\cite{holme2012temporal,harary1997dynamic,casteigts2012time,zhang2017timers}. Many graph-based tasks such as community detection~\cite{lin2008facetnet}, link prediction~\cite{goyal2018dyngem,li2018streaming}, node classification~\cite{jian2018toward}, knowledge graph mining~\cite{trivedi2017know} and network embedding~\cite{li2017attributed, zhou2018dynamic,ma2018depthlgp} haven been shown to be facilitated by including and modeling the temporal information in dynamic graphs. In this work, we propose a dynamic graph neural network model, which incorporates and models the temporal information in the dynamic graphs. 

%% file: sections/conclusions.tex
\section{Conclusion}

In this paper, we propose a novel graph neural graph DGNN for dynamic graphs. It provides two key components -- the update component and the propagation component. When a new edge is introduced, the update component can keep node information being updated by capturing the creation sequential information of edges and the time intervals between interactions.  The propagation component will propagate new interaction information to the influenced nodes by considering influence strengths. We use link prediction and node classification as examples to illustrate how to leverage DGCN to advance graph mining tasks. We conduct experiments on three real-world dynamic graphs and the experimental results in terms of link prediction and node classification suggest the important of dynamic information and the effectiveness of the proposed update and propagation components in capturing dynamic information. 

In the current model, we choose one's neighbors as the set of influenced nodes. Though that choice is reasonable and it works well in practice, we would like to provide some theoretical analysis about this choice and also investigate alternative approaches. Now we illustrate how to use the proposed framework for link prediction and node classification. We also want to investigate how to use the framework for other graph mining tasks especially these under the unsupervised settings such as community detection.